\documentclass{article}
\usepackage{spconf,amsmath,graphicx}
\usepackage{url}
\usepackage{fancyhdr}
\usepackage{textcomp}
\vspace{-.5em}
\vspace{.5em}
\title{Unsupervised Discovery of \textit{Toxoplasma gondii} Motility Phenotypes }
\vspace{1em}
\name{Mojtaba S. Fazli$^{1}$ , Stephen A. Vella$^{2}$, Silvia N.J. Moreno$^{2}$, Shannon Quinn$^{1}$}
\vspace{-1em}
\address{\small{ $^{1}$  Department of Computer Science, The University of Georgia, Athens, GA, USA} \\
 \small{ $^{2}$   Department of Cellular Biology, The University of Georgia, Athens, GA, USA}}
\begin{document}
%\ninept
%
\maketitle
\vspace{-1em}
\begin{abstract}
\vspace{-.5em}
\textit{Toxoplasma gondii} is a parasitic protozoan that causes disseminated toxoplasmosis, a disease that afflicts roughly a third of the world’s population. Its virulence is predicated on its motility and ability to enter and exit nucleated cells; therefore, studies elucidating its mechanism of motility and in particular, its motility patterns in the context of its lytic cycle, are critical to the eventual development of therapeutic strategies. Here, we present an end-to-end computational pipeline for identifying \textit{T. gondii} motility phenotypes in a completely unsupervised, data-driven way. We track the parasites before and after addition of extracellular Ca$^{2+}$ to study its effects on the parasite motility patterns and use this information to parameterize the motion and group it according to similarity of spatiotemporal dynamics.  
\end{abstract}
\begin{keywords}
\textit{Toxoplasma gondii}, computer vision, tracking, spectral clustering, autoregressive models
\end{keywords}
\vspace{-.9em}
\section{Introduction}
\label{sec:intro}
\vspace{-.8em}
\textit{Toxoplasma gondii} is the causative agent of toxoplasmosis and is considered one of the most successful parasitic infections. Approximately 1/3 of the world’s population will test positive for \textit{T. gondii}, a parasite capable of infecting any nucleated cell \cite{saadatnia2012review}. Though \textit{T. gondii} infection is usually mild or asymptomatic within in healthy host, infections are life-long, and the immunocompromised such as AIDS patients, fetuses, and organ-transplant recipients are at risk of developing severe complications from the reactivation of dormant tissue cysts \cite{saadatnia2012review,derouin2008prevention,jones2014neglected}.
Ca$^{2+}$ is a universal signaling molecule that regulates numerous cellular processes in both prokaryotes, unicellular and multicellular eukaryotes \cite{clapham1995calcium}. \textit{T. gondii}’s pathogenesis is directly linked to its lytic cycle, comprised of invasion, replication, egress, and motility, and progression forward is regulated through Ca$^{2+}$ signaling \cite{borges2015calcium}. Therefore, the motion of \textit{T. gondii} is critical to understanding its lytic cycle and, ultimately, in developing potential therapies.  Cytosolic Ca$^{2+}$ oscillations in \textit{T. gondii} precede activation/enhancement of each step of the lytic cycle  \cite{borges2015calcium}, making these oscillations a critical avenue through which to analyze \textit{T. gondii} pathogenesis.
Here we present an unsupervised, data-driven pipeline for identifying \textit{T. gondii} motility patterns in response to cytosolic Ca$^{2+}$. We used extracellular parasites that express in vivo a derivatized variant of the Green Fluorescent Protein (GFP), GCaMP6f, whose fluorescence is directly proportional to Ca$^{2+}$ concentration  \cite{chen2013ultrasensitive}.\\
\newline We linked fluctuations of fluorescence (cytosolic Ca$^{2+}$ concentration) to motility patterns. Using our algorithm, we extracted and quantified motility tracks, parameterized the motion trajectories, and use spectral clustering to identify distinct motility phenotypes. In particular, we repeat this procedure for \textit{T. gondii} parasites both before and after addition of extracellular Ca$^{2+}$, which results in Ca$^{2+}$ influx and stimulation of motility  \cite{borges2015calcium}. This not only allows us to identify distinct \textit{T. gondii} motility phenotypes in a completely unsupervised, data-driven way, but also enriches our knowledge of these phenotypes with respect to the conditions under which they manifest. Ideally, this work will help pave the way to understanding key differences in the biology of \textit{T. gondii} and mammalian cells and the potential for future therapeutic drug discovery.
\vspace{-1.8em}
\section{METHODS}
\vspace{-1.em}
\subsection{Data Acquisition}
\vspace{-.95em}
\label{sec:format}
\textit{T. gondii} tachyzoites (RH) strain were maintained as described previously using Dulbecco’s modified minimal essential media with $1\%$ FBS. GCaMP6f is genetically encoded calcium indicator whose fluorescence is sensitive to cytosolic calcium concentration (Ca$^{2+}$) and allows us to track Ca$^{2+}$ dynamics in motile cells. Construction of GCaMP6f-expressing parasites was performed by cloning the coding sequence of GCaMP6f into a ptubP30GFP (a selection-less plasmid and a kind gift from Boris Striepen). Plasmids were electroporated into RH parasites and subcloned via cell sorting. Videos were acquired using an LSM 710 confocal microscope on $10\%$ FBS (Fetal  Bovine  Serum) coated glass bottom cover dishes (MATTEK)  \cite{borges2015calcium}. Addition of 2µM Thapsigargin and/or $1.8$ mM CaCl2 was noted.  Images were processed using the FIJI ImageJ software suite  \cite{schindelin2012fiji} and converted to z-stacks for processing in OpenCV. For our experiments, we select $10$ videos with average number of $14$ cells in each video.\\
\vspace{-2.5em}
\subsection{Software}
\label{sec:format}
\vspace{-.9em}
We implemented our pipeline using Python 3 and associated scientific computing libraries (NumPy, SciPy, scikit-learn, matplotlib). The core of our tracking algorithm used a combination of tools available in the OpenCV 3.1 computer vision library.  The full code for our pipeline is available under the MIT open source license at \url{https://github.com/quinngroup/toxoplasma-isbi2018} .
\vspace{-1em}
\subsection{Computational Pipeline}
\label{sec:format}

\subsubsection{Tracking Algorithm}

We used a Kanade-Lucas-Tomasi (KLT) tracker to develop trajectories for T. gondii parasites. KLT is an optical flow approach which uses spatial intensities to conduct a focused search of high-probability object locations in the future \cite{lucas1981iterative}\cite{tomasi1991detection}\cite{shi1994good}. It is fast, requiring few matches between sequential images. Here, we identify features extracted from objects in each frame and process them through the KLT module to be tracked over the entire video. The resulting tracker has been shown to be robust in our previous work (Fig. 1) \cite{fazli2017computational}.
\begin{figure}[htb]
  \centerline{\includegraphics[width=8.0cm]{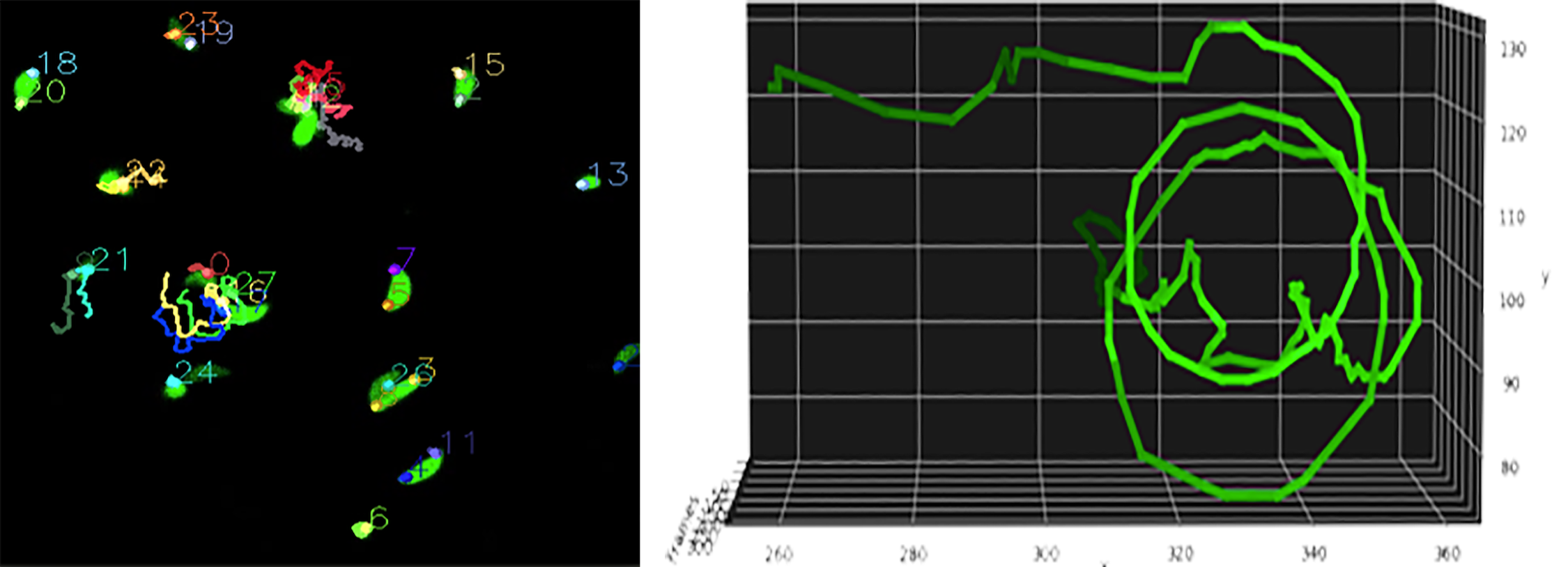}}
\caption{\textit{\textbf{(left)} Object tracking results using KLT.
  \textbf{(right)} 3D visualization of a specific cell’s trajectory. 
  Hue shows the fluorescence of the object, corresponding to Ca$^{2+}$ concentration.
}}
\label{fig:res}
\end{figure}
\vspace{-2em}
\subsubsection{Preprocessing and normalization of KLT trajectories}
KLT trajectories for each parasite object comprised a two-dimensional position vector for each frame. In order to standardize these data to be used in downstream analysis, we employed numerous preprocessing steps. First, we discarded any trajectories below a certain minimum number of frames. This happened if the KLT tracker lost an object (e.g., due to occlusion, movement outside plane of focus). Second, we split the trajectories of each object into two sets, capturing the object trajectory before and after the addition of extracellular Ca$^{2+}$, to be analyzed distinctly. Third, for each trajectory in the two sets, we truncated the number of frames to be equivalent across all objects.

After applying these preprocessing steps, we ended with a corpus containing 139 trajectories, each capturing the two-dimensional movement of distinct \textit{T. gondii} parasites. Each trajectory consisted of 300 frames, half before and half after the addition of extracellular Ca$^{2+}$. Our data were represented in the following structures, X and Y (representing x and y coordinates of each object in each frame, respectively):
\vspace{-.5em}
\[ X_{(139,150)} =
\begin{bmatrix}
    x_{(0,0)} & \dots  & x_{(0,149)} \\
    \vdots & \ddots & \vdots \\
    x_{(138,0)} & \dots  & y_{(138,149)}
\end{bmatrix}
\]
\[ Y_{(139,150)} =
\begin{bmatrix}
    y_{(0,0)} & \dots  & y_{(0,149)} \\
    \vdots & \ddots & \vdots \\
    y_{(138,0)} & \dots  & y_{(138,149)}
\end{bmatrix}
\]
\subsubsection{Parameterization of normalized KLT trajectories}

Performing unsupervised discovery of \textit{T. gondii} motility phenotypes required featurizing the trajectories of each object in such a way as to be invariant to the absolute spatial locations, but sensitive to relative changes over time. Therefore, we used a simple linear model that encodes a Markov-based transition prediction method: an autoregressive (AR) model.
AR models are a type of stochastic process, used heavily in modeling linear systems and predicting future observations as a function of some number of previous ones; in this way, it implements the Markov assumption, in which future observations are independent of all others, conditioned on some number of immediate preceding observations. The number of prior observations used to make future predictions is referred to as the order of the system, and follows as:

\vspace{-.8em}
\[ x_{t} = \sum_{j}^{p} A_{j}x_{t-j} + C\tag{1} \]   

where Eq. 1 denotes a p-order system \cite{10024412700}. In our work, we are interested in the transition matrices $\{A1,..., Ap\}$, which are quantitative encodings of how the system evolves, and therefore represents a parameterization of the system that is not sensitive to absolute spatial positions. 
In our work, we found a system of order $p = 5$ worked well, whereby each $139x150x2$ object was represented as five $2$x$2$ AR motion parameters:$\{A1, A2, A3, A4, A5\}$. These parameters were flattened and concatenated into a single $20$-element feature vector, and this operation was repeated for each of the $139$ objects.
\vspace{-.8em}
\subsubsection{ Establishing pairwise motion similarity}

The AR motion parameters are powerful quantitative representations of movement that are invariant to absolute spatial coordinates. Unfortunately, they are geodesics that do not span a Euclidean space \cite{woolfe2006shift}, and therefore are not amenable to more traditional unsupervised pattern-recognition strategies such as K-means that rely on Euclidean-based pairwise distance metrics.
Instead, we constructed several custom kernel matrices to capture pairwise similarity, with the goal of using these matrices as graph embeddings to use in spectral clustering.
To compute a pairwise similarity (kernel) matrix, we stacked the AR parameters learned from the “before” and “after” sequences of \textit{T. gondii} motility to form two matrices:
\vspace{-.5em}
\[ AR_{Before(139,20)} =
\begin{bmatrix}
    AR_{b(0,0)} & \dots  & AR_{b(0,19)} \\
    \vdots & \ddots & \vdots \\
    AR_{b(138,0)} & \dots  & AR_{b(138,19)}
\end{bmatrix}
\]
\[ AR_{After(139,20)} =
\begin{bmatrix}
    AR_{a(0,0)} & \dots  & AR_{a(0,19)} \\
    \vdots & \ddots & \vdots \\
    AR_{a(138,0)} & \dots  & AR_{a(138,19)}
\end{bmatrix}
\]
We then computed pairwise “affinities,” or similarities, using the radial-basis function (RBF) or Gaussian kernel. This kernel matrix A has the form:

\[ A_{ij}= A_{ji} = e^{-\gamma||x_i - x_j||}\]
where $\gamma$ is a free parameter that designates the width of the Gaussian kernel. Unfortunately, heat kernels are very sensitive to the choice of gamma, so the value was chosen by empirical cross-validation, and selected =0.1 (Fig. 2).
Other kernels were also tested, including trajectory-based kernels such as spatial covariance \cite{wilson2014covariance}, angle covariance (using displacement angles instead of position vectors), and Pearson covariance matrices \cite{weaver2013spss}. However, these did not yield low-dimensional embeddings of sufficient rank to be useful in spectral clustering.
\vspace{-.8em}
\subsubsection{Spectral clustering to identify similar motility phenotypes}
Clustering methods are a natural choice for unsupervised learning; they typically impose few assumptions on the data, defining only a notion of similarity with which to group data together. K-means is a favorite for its speed and simplicity; unfortunately, in our case, the drawbacks of K-means make it an impractical option. For one, K-means assumes Euclidean data; AR motion parameters are geodesics that do not live in a Euclidean space. For two, K-means assumes isotropic clusters; it is possible that AR motion parameters exhibit isotropy in their space, but without a proper distance metric this cannot be explicitly tested.
For these reasons, we chose spectral clustering. Spectral clustering operates on the spectrum of the underlying graph of the data \cite{von2007tutorial}, imposed through a similarity computation using a pairwise kernel. Using this affinity kernel, a graph Laplacian is computed, and its principal eigenvectors are used to embed the original data in a low-dimensional space where they are separable by simpler clustering strategies such as K-means. While this requires a full diagonalization of the Laplacian and can therefore pose a computational bottleneck with large data, our 139 trajectories is sufficiently small for to be solved by out-of-the-box spectral clustering implementations, such as scikit-learn \cite{pedregosa2011scikit}
\begin{figure}[htb]
  \centerline{\includegraphics[width=7.5cm]{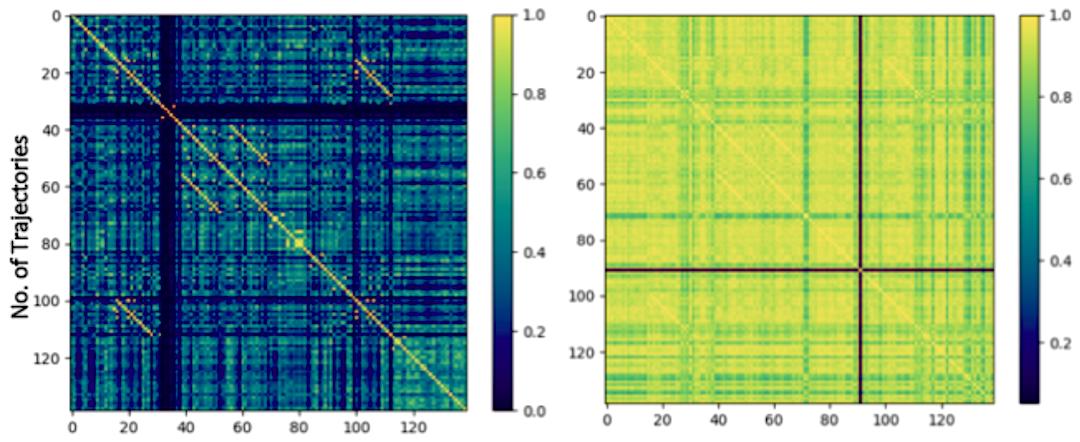}}
\caption{\textit{RBF kernels from AR motion parameters using gamma  = 0.1, from AR motion parameters computed before (left) and after (right) addition of cytosolic Ca$^{2+}$.}}
\label{fig:res}
\end{figure}
\vspace{-1em}
\section{RESULTS AND DISCUSSION}
\vspace{-.8em}
\label{sec:pagestyle}
Using established heuristics from previous work on the \textit{T. gondii} parasite and its lytic cycle \cite{borges2015calcium, fazli2017computational}, we chose k = 5 clusters for our analysis.
Figure 3 shows a summary of the clustering results. The left column indicates trajectories from before the addition of cytosolic Ca$^{2+}$, and the right column of after Ca$^{2+}$ addition. Each subfigure is a single object, illustrated by a line showing the extent of its movement over the 150-frame duration of the segment.

First, we observe that object trajectories are significantly richer and more dynamic after the addition of cytosolic Ca$^{2+}$ than before; we saw this in Fig. 1 (right), and is well-understood that since Ca$^{2+}$ regulates the \textit{T. gondii} lytic cycle, we can expect more activity when those concentrations are high \cite{clapham1995calcium,borges2015calcium}. 
Second, and most relevant to the goal of this work, is that we can observe qualitatively unmistakable similarity of trajectory dynamics, especially in the after-Ca$^{2+}$ clusters. Most of the before-Ca$^{2+}$ trajectories are largely immotile (Fig. 3, bottom left), but there are clusters that show similar motion motifs, including full or partial circles (Fig. 3, upper left). The after-Ca$^{2+}$ clusters show similar groupings of motifs: a clear partial-twirl, sometimes combined with a straight trajectory (Fig. 3, lower right), or seemingly random motion with a long outward arc (Fig. 3, upper right).
\begin{figure}[htb]
  \centerline{\includegraphics[width=8.5cm]{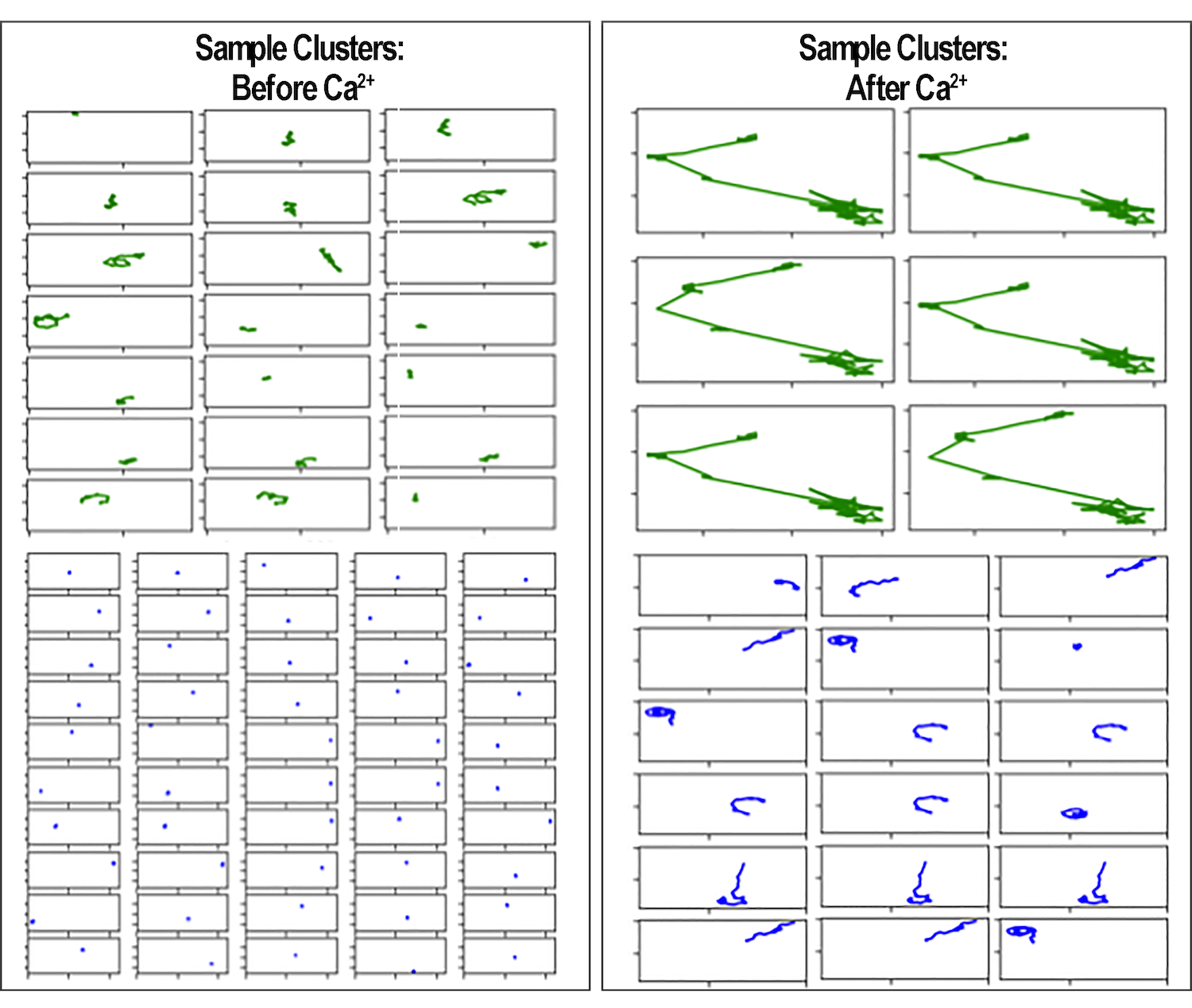}}
\caption{\textit{Clustering results of trajectories before \textbf{(left column)} and after \textbf{(right column)} addition of cytosolic Ca$^{2+}$. Two sample clusters are shown in each column. Subplots are aligned to center for easy viewing.
}}
\label{fig:res}
\end{figure}
Particularly when viewed in context (Fig. 4), we can clearly identify several distinct \textit{T. gondii} motility phenotypes— such as helical, twirling, and circular—that are identified through manual heuristics [1-5].

There are limitations to this approach. First, while the RBF kernel encodes nonlinear relationships, it still relies on a Euclidean-style pairwise comparison. Second, despite the empirical performance of AR parameters in encoding two-dimensional spatial movement, this is not an ideal application of this technique; it is likely that the relative simplicity of \textit{T. gondii} motility patterns, and the small number of “motion motifs,” are responsible for the relatively good empirical performance. Third, the combination of small quantity of data and low-resolution temporal models precludes a deeper exploration of the manifold of \textit{T. gondii} motility phenotypes. Nevertheless, the results strongly suggest such a manifold exists, and could be elucidated in future work.
\vspace{-.8em}
\begin{figure}[htb]
  \centerline{\includegraphics[width=8.5cm]{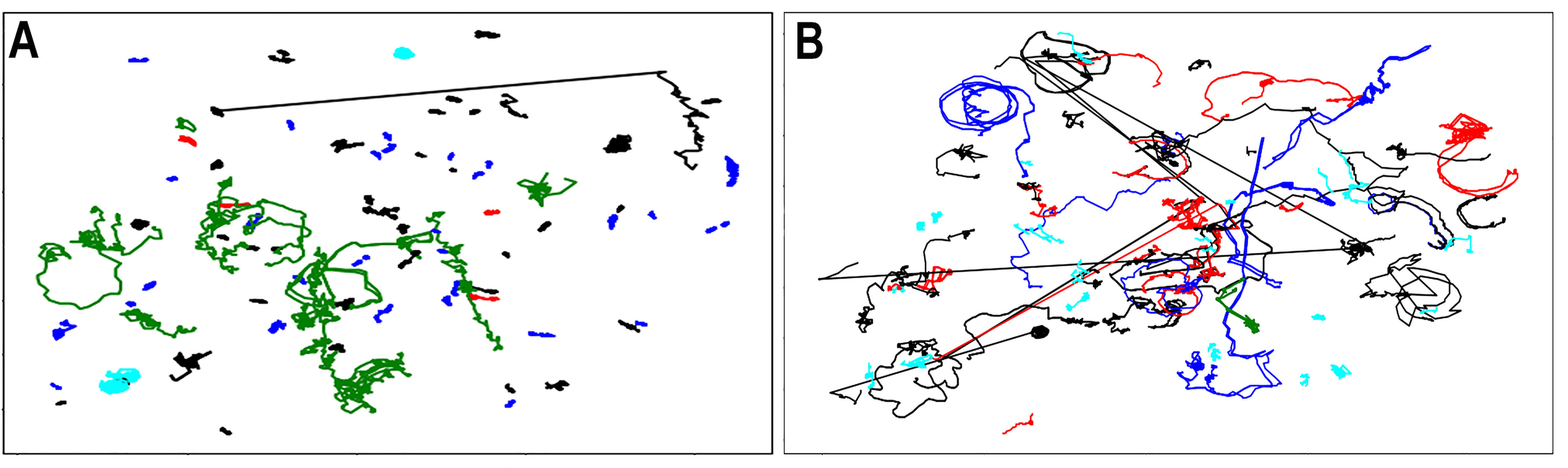}}
\caption{\textit{Aggregated plotting of trajectories before (A) and after (B) addition of cytosolic Ca$^{2+}$. Each object is colored by its cluster identifier, and placed according to its original absolute spatial location. Here, some trajectories(straight lines mostly in part B) are artifacts and were excluded from the final clustering results.
}}
\label{fig:res}
\end{figure}
\vspace{-2em}
\section{conclusion}
\vspace{-1em}
\label{sec:typestyle}
In this work, we have demonstrated a computational pipeline for tracking, extracting, parameterizing, and clustering motion trajectories of \textit{T. gondii} parasites in an initial effort to elucidate discrete patterns in the parasite’s lytic cycle. This is a first step toward a mechanistic understanding of the parasite’s virulence as a function of its motility patterns. We have also demonstrated the application of a linear time series (AR) model to the parameterization of two-dimensional spatial motility, and its effectiveness in clustering for recognizing gross motion patterns. Given the relatively short tracks of the \textit{T. gondii} objects, we felt AR models (i.e., a linear model) would be sufficient to parameterize motion. This technique could be applied pretty much as-is to any trajectory analysis of discrete objects.

In the future, we anticipate refining the elements of this pipeline to extract finer-grained motility patterns. Specifically, we anticipate deep learning to be the next evolution of this pipeline: using segmentation and tracking tools such as Faster RCNN to track the objects, and parameterizing the motility sequences using recurrent neural networks instead of linear systems.
\vspace{-.8em}
\section{ACKNOWLEDGMENTS}
\vspace{-.8em}
\label{sec:majhead}

The authors acknowledge Zhu Hong Li for construction of the GCaMP6f non-selectable strain. This work was supported in part by AWS in Education Grant Award. We acknowledge partial support from the NSF Advances in Biological Informatics (ABI) under award number 1458766.The Toxoplasma work was funded by the National Institutes of Health (AI-110027 and AI-096836 to SNJM).
\vspace{-1em}
\small{
\bibliographystyle{IEEEbib}
\bibliography{strings,refs}
}
\end{document}